\documentclass{article}
\usepackage{ijcai21}

\usepackage{times}
\usepackage{soul}
\usepackage{url}
\usepackage[hidelinks]{hyperref}
\usepackage[utf8]{inputenc}
\usepackage[small]{caption}
\usepackage{graphicx}
\usepackage{amsmath}
\usepackage{amsthm}
\usepackage{booktabs}
\usepackage{algorithm}
\usepackage{algorithmic}
\usepackage{multirow}
\usepackage{latexsym}
\usepackage{float}
\usepackage{colortbl}
\usepackage{amsmath,amssymb,amsfonts}
\usepackage[mathscr]{euscript}

\def\ie{\emph{i.e.}}
\def\eg{\emph{e.g.}}

\newcommand{\secref}[1]{$\S$ \ref{#1}}
\def\OurModel{\textit{C$^2$F-Net}} 

\graphicspath{{./Imgs/}}
\DeclareGraphicsExtensions{.jpg,.pdf,.png}

\title{Context-aware Cross-level Fusion Network for Camouflaged Object Detection }

\author{
Yujia Sun$^{1}$\footnote{Co-first Authors.}\and
Geng Chen$^{2*}$\and
Tao Zhou$^{3}$\footnote{Corresponding Author (taozhou.ai@gmail.com).}\and
Yi Zhang$^4$\and
Nian Liu$^2$\\ 
\affiliations
$^1$School of Computer Science, Inner Mongolia University, China \quad
$^2$ IIAI \\
$^3$PCA Lab, the Key Laboratory of Intelligent Perception and Systems for High-Dimensional Information of Ministry of Education, School of Computer Science and Engineering, Nanjing University of Science and Technology, China \quad
$^4$Institut National des Sciences Appliquées de Rennes \\ 
}

\begin{document}

\maketitle

\begin{abstract}
Camouflaged object detection (COD) is a challenging task due to the low boundary contrast between the object and its surroundings.
In addition, the appearance of camouflaged objects varies significantly, \eg, object size and shape, aggravating the difficulties of accurate COD.
In this paper, we propose a novel Context-aware Cross-level Fusion Network (\OurModel) to address the challenging COD task.
Specifically, we propose an Attention-induced Cross-level Fusion Module (ACFM) to integrate the multi-level features with informative attention coefficients.
The fused features are then fed to the proposed Dual-branch Global Context Module (DGCM), which yields multi-scale feature representations for exploiting rich global context information.
In \OurModel, the two modules are conducted on high-level features using a cascaded manner.
Extensive experiments on three widely used benchmark datasets demonstrate that our \OurModel~is an effective COD model and outperforms state-of-the-art models remarkably. Our code is publicly available at: \href{https://github.com/thograce/C2FNet}{https://github.com/thograce/C2FNet}.
\end{abstract}

\section{Introduction}
\begin{figure}[t]
    \centering{
    }
	\includegraphics[width=1\linewidth]{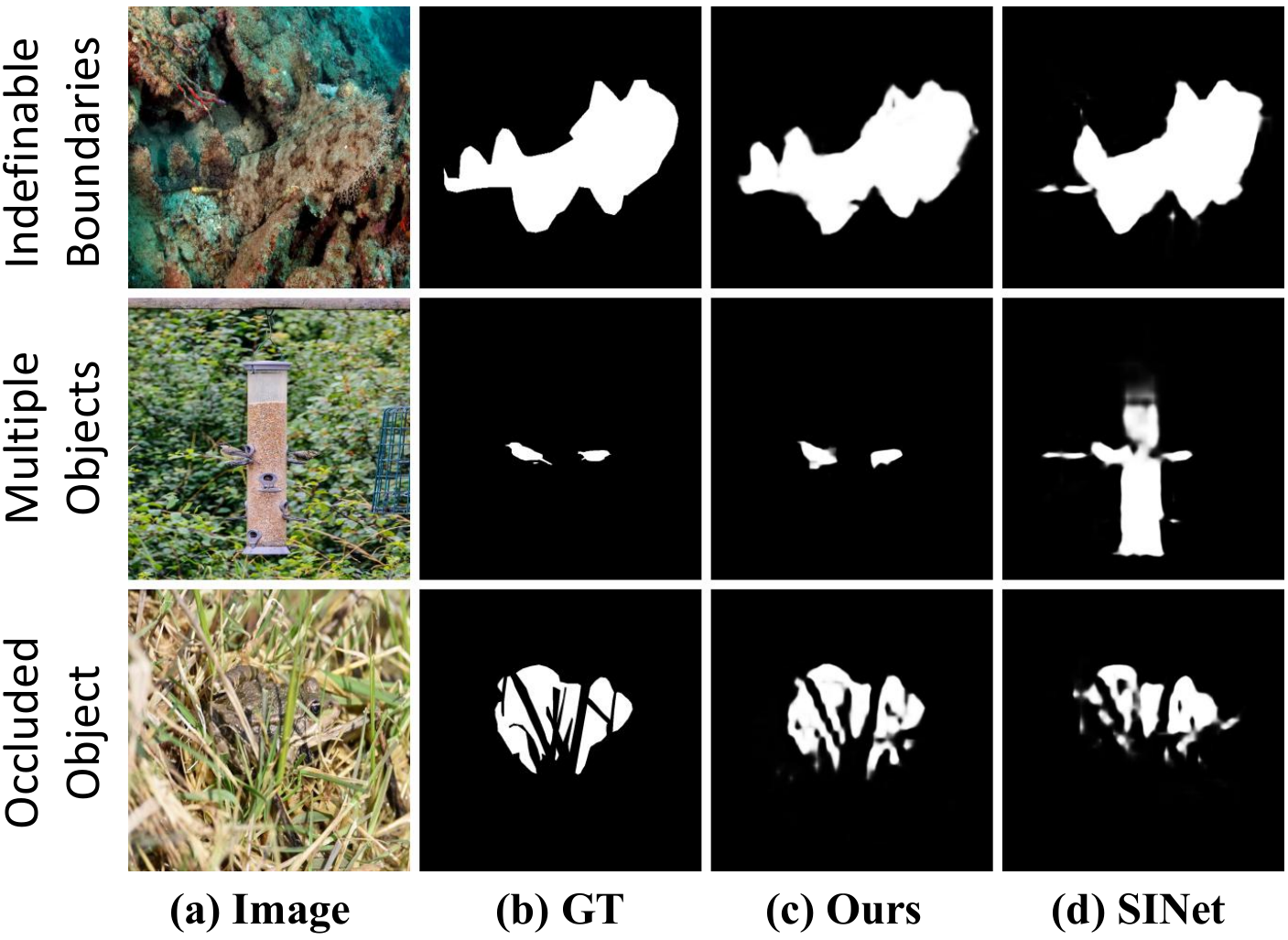}
	\caption{From top to bottom, three challenging camouflage scenarios with indefinable boundaries, multiple objects and occluded object are listed. Our model outperforms SINet \protect\cite{fan2021concealed} under these challenging scenarios.}
    \label{fig:examples}
\end{figure}

Due to the purpose of survive, wild animals have developed rich camouflage abilities.
In practice, they try to change their appearance to ``perfectly'' integrate into the surrounding environment, in order to avoid the attention from other creatures.
In recent years, camouflage has attracted increasing research interest from the computer vision community \cite{anet,fan2021concealed,Lyu2021,mei2021Ming,zhai2021Mutual}.
Among various topics, camouflaged object detection (COD), which aims to identify and segment the camouflaged objects from images, is particularly popular.
However, COD is a challenging task due to the nature of camouflage. More specifically, due to the camouflage, the boundary contrast between an object and its surroundings is extremely low, leading to significant difficulties to identify/segment this object. As shown in the top row of Figure~\ref{fig:examples}, it is very challenging to discover the fish from the underwater image.
In addition, the camouflaged objects, most wild animals, are usually with varied appearance, \eg, size and shape, which further aggravates the difficulties of accurate COD \cite{anet,fan2021concealed}.

To address these challenges, deep learning techniques have been adopted and shown great potentials \cite{anet,fan2021concealed}.
In addition, a number of COD datasets were constructed for training the deep learning models.
For instance, \cite{anet} created the first COD dataset, called CAMO, consisting of 2500 images.
However, due to the limited sample size, CAMO is unable to take full advantage of deep learning models.
Recently, \cite{fan2021concealed} constructed COD10K, the first large-scale COD dataset containing 10,000 images with the consideration of various challenging camouflage attributes in the real natural environments. Apart from datasets, these two works also contribute to the COD from the aspect of deep learning models. \cite{anet} proposed an anabranch network, which performs classification to predict whether an image contains camouflaged objects, and then integrates this information into the camouflaged object segmentation task. Based on the observation that predators discover preys by first searching and then recognizing, \cite{fan2021concealed} designed a searching module and an identification module to locate rough areas and to accurately segment camouflaged objects, respectively.

Existing methods have shown promising performance in the detection of a single camouflaged object from relatively simple scenes. However, their performance degrades significantly for a number of challenging cases, such as multiple objects, occlusion, etc.
As shown in Figure~\ref{fig:examples}, stat-of-the-art model fails to correctly identify all camouflaged objects and show unsatisfactory performance when occlusion exits.
A key solution to address these challenges is a significant large receptive field, which provides rich context information for accurate COD.
In addition, the cross-level feature fusion also plays a vital role in the success of COD.
However, existing works usually overlook the importance of these two key factors. It is, therefore, greatly desired for a unified COD framework that jointly considers both rich context information and effective cross-level feature fusion.

To this end, we propose a novel COD model, called Context-aware Cross-level Fusion Network (\OurModel).
We first fuse the cross-level features extracted from the backbone using an Attention-induced Cross-level Fusion Module (ACFM), which integrates the features with the guidance from a Multi-Scale Channel Attention (MSCA) component.
The ACFM computes the attention coefficients from multi-level features, employs the resulting attention coefficients to refine the features, and then integrates the resulting features for the fused one.
Subsequently, we propose a Dual-branch Global Context Module (DGCM) to exploit the rich context information from the fused features.
The DGCM transforms the input feature into multi-scaled ones with two parallel branches, employs the MSCA components to compute the attention coefficients, and integrates the features by considering the attention coefficients.
The ACFM and DGCM are organized in a cascaded manner at two stages, from high-level to low-level.
The final DGCM predicts the segmentation map of the camouflaged object(s) from an image.
In summary, the main contributions of this paper are as follows:
\begin{itemize}
    
    \item We propose a novel COD model, \OurModel, which integrates the cross-level features with the consideration of rich global context information.
    
    \vspace{-5pt}
    \item We propose a context-aware module, DGCM, which exploits global context information from the fused features. Our DGCM is capable of capturing valuable context information, which is the key factor for improving COD accuracy. 
    
    \vspace{-5pt}
    \item We integrate the cross-level features with an effective fusion module, ACFM, which integrates the features with the valuable attention cues provided by MSCA. 
    
    \vspace{-5pt}
    \item Extensive experiments on three benchmark datasets demonstrate that our \OurModel~outperforms 14 state-of-the-art models in the terms of four evaluation metrics.
\end{itemize}

\section{Related Work}

In this section, we discuss the two types of works that are closely related to our model, \ie, camouflaged object detection and context-aware deep learning.

\subsection{Camouflaged Object Detection}
As an emerging field, COD has gained increasing attention in recent years.
Early works rely on visual features, such as color, texture, motion, and gradient~\cite{survey}.
In practice, the techniques relying on a single feature are unable to provide promising performance, therefore \cite{comb1} integrated different features for improved performance.
However, limited by hand-crafted features, these methods are only suitable for relatively simple scenarios and tend to fail in real-world applications.
To address this, deep learning techniques have been employed for COD by taking advantage of deep features automatically learned by the network from extensive training images, which are more generic and effective than hand-crafted features.
For instance, \cite{cdetection} identified the camouflaged object with a region proposal network. \cite{mirrornet} observed that flipping the image provides vital cues for identifying the camouflaged object, therefore a two-stream MirrorNet was proposed with the original and flipped images as inputs for COD.
\cite{video} exploited the motion information to identify camouflaged objects in videos with a deep learning framework, consisting of two components, \ie, a differentiable registration module to highlight object boundaries and a motion segmentation module with memory that discovers moving regions.
This method overcomes the limitations of traditional motion information based methods that tend to fail when the target is stationary. For COD application, polyp segmentation has also received widespread attention. \cite{pranet} proposed a parallel reverse attention network (PraNet), which first predicts rough regions and then refines the boundaries.

\begin{figure*}[t!]
    \centering{
    }
	\includegraphics[width=.9\textwidth]{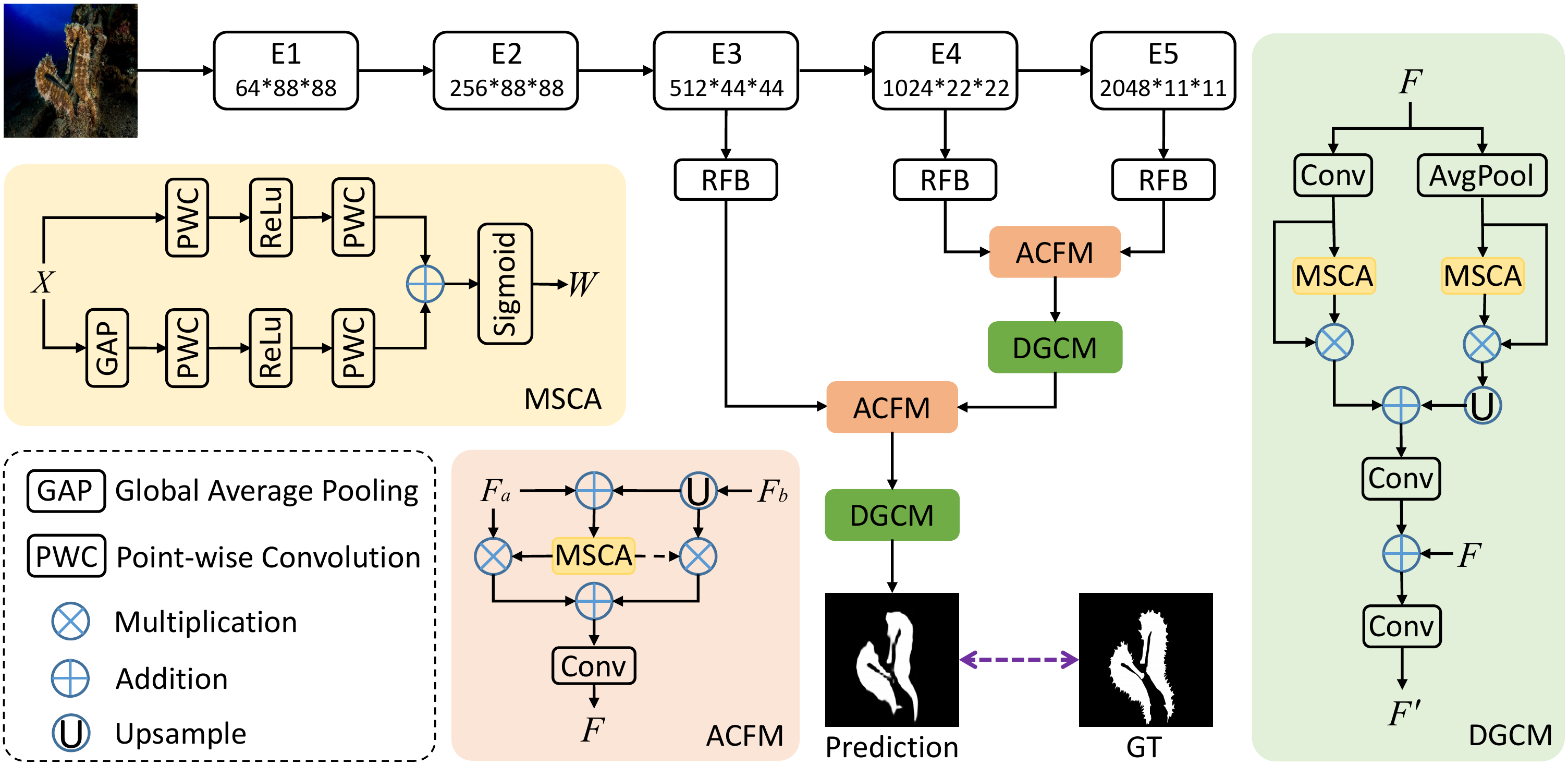}
	\caption{The overall architecture of the proposed model, which consists of two key components, \ie, attention-induced cross-level fusion module and dual-branch global context module. See \secref{Methods} for details.
	}
    \label{fig:net}
\end{figure*}

\subsection{Context-aware Deep Learning}
The contextual information plays a crucial role in object segmentation tasks due to its great capability of enhancing the feature representation for improved performance.
Efforts have been dedicated to enriching the contextual information.
For instance, PSPNet~\cite{pspnet} obtains rich context information by establishing multi-scale representations around each pixel. \cite{deeplab} employed different dilated convolutions to construct ASPP to achieve context information acquisition. DANet~\cite{danet} utilizes non-local modules~\cite{nonlocal} to extract contextual information. CCNet~\cite{ccnet} obtains dense contextual information by cyclically cross-attention modules.
In the field of salient object detection, \cite{bi} utilized a multi-scale context-aware feature extraction module to allow multi-level feature maps to obtain rich context information. PoolNet~\cite{poolnet}, which is equipped with a pyramid pool module, employs a global guidance module to obtain the location information of underlying salient targets for different feature layers. GCFANet~\cite{gcpanet} utilizes a global context flow module to transfer features containing global semantics to feature mapping at different stages to improve the integrity of the detection of prominent objects. LCANet~\cite{lcanet} adaptively integrates the local area context and the global scene context in a coarse-to-fine structure to enhance local context features.

\section{Proposed Method}
\label{Methods}

In this section, we first outline the overall architecture of the proposed \OurModel. Then, we provide the details of the two proposed modules. Finally, we give the overall loss function for the proposed model.


\subsection{Overall Architecture}

Figure~\ref{fig:net} shows the overall architecture of the proposed \OurModel, which fuses context-aware cross-level features to improve the camouflaged object detection performance. Specifically, we adopt Res2Net-50~\cite{res2net} to extract its  features at five different layers, denoted as $f_i~(i=1,2,\dots,5)$.
Then, a receptive field block (RFB) is adopted to expand the receptive field for capturing richer features in a specific layer. We utilize the same setting in \cite{res2net}, and the RFB component includes five branches $b_k~(k=1,2,\dots,5)$. In each branch, the first convolutional layer has dimensions $1\times{1}$ to reduce the channel size to 64. This is followed by two layers, \ie, a $(2k-1)\times(2k-1)$ convolutional layer and a $3\times{3}$ convolutional layer with a specific dilation rate $(2k-1)$ when $k>2$. The first four branches are concatenated and then their channel size is reduced to $64$ using a $1\times{1}$ convolutional operation. Then, the fifth branch is added and the whole module is fed to a \emph{ReLU} activation function to obtain the final feature. After that, we propose an Attention-induced Cross-level Fusion Module (ACFM) to integrate multi-scale features, and a Dual-branch Global Context Module (DGCM) to mine multi-scale context information within the fused features. Finally, we obtain the prediction result for camouflaged object detection. We give the details of each key component below.



\subsection{Attention-induced Cross-level Fusion Module}

There are natural differences among different types of camouflaged objects. Besides, the size of similar camouflaged objects may also vary greatly, due to the observation distance and their relative location to the surrounding environment. In other words, for a captured image with camouflaged object(s), the proportion of the camouflaged object(s) often varies. To address these challenges, we propose an ACFM via introducing a Multi-Scale Channel Attention (MSCA)~\cite{aff} mechanism to effectively fuse cross-level features, which can exploit multi-scale information to alleviate scale changes. The MSCA has strong adaptability to different scale targets, and its structure is shown in Figure~\ref{fig:net}. MSCA is based on a double-branch structure, in which one branch uses global average pooling to obtain global contexts to emphasize globally distributed large objects, and the other branch maintains the original feature size to obtain local contexts to avoid small objects being ignored. Different from other multi-scale attention mechanisms, MSCA uses point convolution in two branches to compress and restore features along the channel dimension, thereby aggregating multi-scale channel context information. Features at different levels have different contributions to the task, and fusion of features at multiple levels can complement each other to obtain a comprehensive feature expression. 

As discussed in \cite{cpd}, low-level features with greater spatial resolution demand more computing resources than high-level features, but contribute less to the performance of deep integration models. Motivated by this observation, we only carry out ACFM in high-level features. Specifically, we call ${f_{i}~(i=3, 4, 5)}$ as high-level features, and the cross-level fusion process can be described as follows:
\begin{equation}
\resizebox{.91\linewidth}{!}{$
    \displaystyle
    F_{ab}=\mathcal{M}(F_{a}\uplus F_{b})\otimes F_{a}\oplus(1-\mathcal{M}(F_{a}\uplus F_{b}))\otimes F_{b},
$}
\end{equation}%
where $\mathcal{M}$ represents the MSCA, $F_{a}$ and $F_{b}$ represent the two input features. $\oplus$ represents the initial fusion of $F_{a}$ and $F_{b}$, that is, $F_{b}$ is up-sampled twice and then added element-wisely with $F_{a}$. $\otimes$ represents element-wise multiplication. And $(1-\mathcal{M}(F_{a}\uplus F_{b}))$ corresponds to the dotted line in Figure~\ref{fig:net}. After that, the feature $F_{ab}$ is fed into a $3\times{3}$ convolutional layer, followed by batch normalization and a \emph{ReLU} activation function. Finally, we obtain the cross-level fused feature $F$.

\subsection{Dual-branch Global Context Module}

We have adopted the proposed ACFM to fuse multi-scale features at different levels, which introduces the multi-scale channel attention mechanism to obtain informative attention-based fused features. Besides, global context information is critical for improving the camouflaged object detection performance. Thus, we propose a Dual-branch Global Context Module (DGCM) to exploit rich global context information within the fused cross-level features. Specifically, the output $F\in\mathbb{R}^{{W}\times{H}\times{C}}$ of the ACFM is fed to two branches though a convolutional operation and average pooling, respectively. We can obtain the sub features $F_{c}\in\mathbb{R}^{{W}\times{H}\times{C}}$ and $F_{p}\in\mathbb{R}^{{\frac{W}{2}}\times{\frac{H}{2}}\times{C}}$. To learn multi-scale attention-based feature representations, $F_{c}$ and $F_{p}$ are first fed to the MSCA component. Element-wise multiplication is adopted to fuse the outputs of MSCA and the corresponding features ($F_{c}$ or $F_{p}$), and then we obtain $F_{cm}\in\mathbb{R}^{{W}\times{H}\times{C}}$ and $F_{pm}\in\mathbb{R}^{{\frac{W}{2}}\times{\frac{H}{2}}\times{C}}$. After that, we directly fuse the features from the two branches using an addition operation to obtain $F_{cpm}$. Finally, a residual structure is used to fuse $F$ and $F_{cpm}$ to get ${F}'$. The above process can be described as follows:
\begin{equation}
\left\{
\begin{aligned}
&F_{c}= \mathcal{C}\left ( F \right ), F_{cm}= F_{c}\otimes \mathcal{M}\left ( F_{c} \right ),\\
&F_{p}= \mathcal{C}\left( \mathcal{P} \left ( F \right )\right), F_{pm}= F_{p}\otimes \mathcal{M} \left ( F_{p} \right ),\\
&{F}'= \mathcal{C} \left ( F\oplus \mathcal {C} \left ( F_{cm} \oplus \mathcal{U} \left ( F_{pm} \right )\right ) \right ),
\end{aligned}
\right.
\label{eq01}
\end{equation}
where $\mathcal{C}$, $\mathcal{P}$, $\mathcal{M}$, and $\mathcal{U}$ represent the convolution, average pooling, MSCA, and upsampling operations, respectively.
It is worth noting that the proposed DGCM is used to enhance the fusion features of ACFM, which makes our model can adaptively extract multi-scale information from a specific level during the training phase. The detailed structure is shown in Figure~\ref{fig:net}.

\subsection{Loss Function}
The binary cross-entropy loss ($\mathcal{L}_\text{BCE}$) is used to independently calculate the loss of each pixel to form a pixel restriction on the network. In order to make up for its deficiency of ignoring the global structure, \cite{basnet} introduced IoU loss($\mathcal{L}_\text{IoU}$) to focus on the global structure to form a global restriction on the network. These losses are treated equally to all pixels, ignoring the differences between pixels. Based on this, \cite{f3net} improved the above two losses into the weighted binary cross-entropy loss~($\mathcal{L}_\text{BCE}^w$) and IoU loss ($\mathcal{L}_\text{IoU}^w$). By calculating the difference between the center pixel and its surrounding environment, each pixel is assigned a different weight, so that the hard pixels can be obtained more attention. In summary, the loss function of our model is defined as: $\mathcal{L} = \mathcal{L}_\text{IoU}^w + \mathcal{L}_\text{BCE}^w$. The network prediction result $P$ is up-sampled to have the same size with ground-truth map $G$. Thus, the total loss for the proposed model can be formulated as: $\mathcal{L}_{\text{total}} = \mathcal{L} (G, P)$.

\section{Experiments}

In this section, we first provide the implementation details, datasets and evaluation metrics used in our experiments. Then, we provide quantitative and qualitative comparison results. Finally, we present the results of ablation studies to validate the effectiveness of key modules.

\subsection{Implementation Details}

We implement our model using PyTorch and adopt Res2Net-50~\cite{res2net} pretrained on ImageNet as our network backbone. We carry out a multi-scale training strategy $\{0.75, 1, 1.25\}$ instead of data augmentation and resize all input images to $352\times{352}$. We utilize the AdaX~\cite{li2020adax} optimization algorithm to optimize the overall parameters by setting the initial learning rate as $1e-4$. Accelerated by an NVIDIA Tesla P40 GPU, the whole network takes about 4 hours to converge over 40 epochs for training with a batch size of 32. During training, the learning rate decays 10 times after 30 epochs.

\subsection{Datasets}
We conduct experiments on three public benchmark datasets for camouflaged object detection. The details of each dataset are as follows:

\begin{table*}[ht]
\caption{Quantitative comparison with state-of-the-art methods for COD on three benchmark datasets using four widely used evaluation metrics (\ie, $S_{\alpha}$, $E_{\phi}$, $F_{\beta}^w$, and $M$). ``$\uparrow$" / ``$\downarrow$" indicates that larger or smaller is better. The best results are highlighted in \textbf{Bold} fonts.}\vspace{-0.25cm}
\renewcommand{\arraystretch}{0.98}
\setlength\tabcolsep{6.5pt}
\footnotesize
\begin{tabular}{l|l|llll|llll|llll}
\hline\toprule
\multicolumn{1}{l|}{\multirow{2}{*}{Method}} & 
\multicolumn{1}{c|}{\multirow{2}{*}{Year}} & \multicolumn{4}{c|}{CHAMELEON} & \multicolumn{4}{c|}{CAMO-Test} & \multicolumn{4}{c}{COD10K-Test} \\ \cline{3-14}
 & \multicolumn{1}{c|}{} & \multicolumn{1}{c}{$S_\alpha\uparrow$} & \multicolumn{1}{c}{$E_\phi\uparrow$} & \multicolumn{1}{c}{$F_\beta^w\uparrow$} & \multicolumn{1}{c|}{$M\downarrow$} & \multicolumn{1}{c}{$S_\alpha\uparrow$} & \multicolumn{1}{c}{$E_\phi\uparrow$} & \multicolumn{1}{c}{$F_\beta^w\uparrow$} & \multicolumn{1}{c|}{$M\downarrow$} & \multicolumn{1}{c}{$S_\alpha\uparrow$} & \multicolumn{1}{c}{$E_\phi\uparrow$} & \multicolumn{1}{c}{$F_\beta^w\uparrow$} & \multicolumn{1}{c}{$M\downarrow$} \\ \midrule
FPN & 2017 & 0.794 & 0.783 & 0.590 & 0.075 & 0.684 & 0.677 & 0.483 & 0.131 & 0.697 & 0.691 & 0.411 & 0.075 \\ 
MaskRCNN & 2017 & 0.643 & 0.778 & 0.518 & 0.099 & 0.574 & 0.715 & 0.430 & 0.151 & 0.613 & 0.748 & 0.402 & 0.080 \\ 
PSPNet & 2017 & 0.773 & 0.758 & 0.555 & 0.085 & 0.663 & 0.659 & 0.455 & 0.139 & 0.678 & 0.680 & 0.377 & 0.080 \\ 
UNet++ & 2018 & 0.695 & 0.762 & 0.501 & 0.094 & 0.599 & 0.653 & 0.392 & 0.149 & 0.623 & 0.672 & 0.350 & 0.086 \\ 
PiCANet & 2018 & 0.769 & 0.749 & 0.536 & 0.085 & 0.609 & 0.584 & 0.356 & 0.156 & 0.649 & 0.643 & 0.322 & 0.090 \\ \midrule
MSRCNN & 2019 & 0.637 & 0.686 & 0.443 & 0.091 & 0.617 & 0.669 & 0.454 & 0.133 & 0.641 & 0.706 & 0.419 & 0.073 \\ 
BASNet & 2019 & 0.687 & 0.721 & 0.474 & 0.118 & 0.618 & 0.661 & 0.413 & 0.159 & 0.634 & 0.678 & 0.365 & 0.105 \\ 
PFANet & 2019 & 0.679 & 0.648 & 0.378 & 0.144 & 0.659 & 0.622 & 0.391 & 0.172 & 0.636 & 0.618 & 0.286 & 0.128 \\ 
CPD & 2019 & 0.853 & 0.866 & 0.706 & 0.052 & 0.726 & 0.729 & 0.550 & 0.115 & 0.747 & 0.770 & 0.508 & 0.059 \\ 
HTC & 2019 & 0.517 & 0.489 & 0.204 & 0.129 & 0.476 & 0.442 & 0.174 & 0.172 & 0.548 & 0.520 & 0.221 & 0.088 \\ 
EGNet & 2019 & 0.848 & 0.870 & 0.702 & 0.050 & 0.732 & 0.768 & 0.583 & 0.104 & 0.737 & 0.779 & 0.509 & 0.056 \\ 
ANet-SRM & 2019 & \multicolumn{1}{c}{‡} & \multicolumn{1}{c}{‡} & \multicolumn{1}{c}{‡} & \multicolumn{1}{c|}{‡} & 0.682 & 0.685 & 0.484 & 0.126 & \multicolumn{1}{c}{‡} & \multicolumn{1}{c}{‡} & \multicolumn{1}{c}{‡} & \multicolumn{1}{c}{‡} \\ \midrule
SINet & 2020 & 0.869 & 0.891 & 0.740 & 0.044 & 0.751 & 0.771 & 0.606 & 0.100 & 0.771 & 0.806 & 0.551 & 0.051 \\ 
MirrorNet & 2020 & \multicolumn{1}{c}{‡} & \multicolumn{1}{c}{‡} & \multicolumn{1}{c}{‡} & \multicolumn{1}{c|}{‡} & 0.741 & 0.804 & 0.652 & 0.100 & \multicolumn{1}{c}{‡} & \multicolumn{1}{c}{‡} & \multicolumn{1}{c}{‡} & \multicolumn{1}{c}{‡} \\ 
\OurModel~(Ours) & 2021 & \textbf{0.888} & \textbf{0.935} & \textbf{0.828} & \textbf{0.032} & \textbf{0.796} & \textbf{0.854} & \textbf{0.719} & \textbf{0.080} & \textbf{0.813} & \textbf{0.890} & \textbf{0.686} & \textbf{0.036} \\ 
\bottomrule
\hline
\end{tabular}
\label{tab1}
\end{table*}

\begin{itemize}
    \item CHAMELEON~\cite{fan2021concealed} contains 76 camouflaged images, which are collected through the Google search engine with the keyword ``camouflage animals".
    
    \item CAMO~\cite{anet} has 1,25k images (1k for training, 0.25k for testing), including 8 categories.
    
    \item COD10K~\cite{fan2021concealed} is currently the largest camouflaged object detection dataset. It consists of 5,066 camouflaged images (3040 for training, 2026 for testing), which is divided into 5 super-classes and 69 sub-classes. This dataset also provides high-quality fine annotation, reaching the level of matting.
\end{itemize}


\textbf{Training/Testing Sets}.
Following the training setting in~\cite{fan2021concealed}, for both CAMO and COD10K datasets, we use the default training sets.
Then, we evaluate our model and all compared methods on the whole CHAMELEON dataset and the test sets of CAMO and COD10K.

\begin{table*}[h]
\caption{Quantitative evaluation for ablation studies on the three datasets. The best results are highlighted in \textbf{Bold} fonts. Ver. = Version.} \vspace{-0.25cm}
\renewcommand{\arraystretch}{0.75}
\setlength\tabcolsep{5.6pt}
\footnotesize
\begin{tabular}{l|l|llll|llll|llll}
\hline\toprule
\multirow{2}{*}{Ver.} & \multicolumn{1}{l|}{\multirow{2}{*}{Method}} & \multicolumn{4}{c|}{CHAMELEON} & \multicolumn{4}{c|}{CAMO-Test} & \multicolumn{4}{c}{COD10K-Test} \\ \cline{3-14} 
 & \multicolumn{1}{c|}{} & \multicolumn{1}{c}{$S_\alpha\uparrow$} & \multicolumn{1}{c}{$E_\phi\uparrow$} & \multicolumn{1}{c}{$F_\beta^w\uparrow$} & \multicolumn{1}{c|}{$M\downarrow$} & \multicolumn{1}{c}{$S_\alpha\uparrow$} & \multicolumn{1}{c}{$E_\phi\uparrow$} & \multicolumn{1}{c}{$F_\beta^w\uparrow$} & \multicolumn{1}{c|}{$M\downarrow$} & \multicolumn{1}{c}{$S_\alpha\uparrow$} & \multicolumn{1}{c}{$E_\phi\uparrow$} & \multicolumn{1}{c}{$F_\beta^w\uparrow$} & \multicolumn{1}{c}{$M\downarrow$} \\ \midrule
No.1 & Basic &0.882  &0.928  &0.813  &0.033  &0.774  &0.829  &0.684  &0.090  &0.805  &0.880  &0.668  &0.038  \\ \midrule
No.2 & Basic+ACFM &0.886 &\textbf{0.944}  &0.823  &\textbf{0.028}  &0.784  &0.831  &0.694  &0.086  &0.808  &0.881  &0.675  &0.037  \\ \midrule
No.3 & Basic+DGCM &0.883 &0.936  &0.820  &0.030  &0.786  &0.840  &0.696  &0.081  &0.808  &0.881  &0.676  &\textbf{0.036}  \\ \midrule
No.4 & Basic+ACFM+DGCM &\textbf{0.888}  &0.935  &\textbf{0.828}  &{0.032}  &\textbf{0.796}  &\textbf{0.854}  &\textbf{0.719}  &\textbf{0.080}  &\textbf{0.813}  &\textbf{0.890}  &\textbf{0.686}  &\textbf{0.036}  \\ \bottomrule
\hline
\end{tabular}
\label{tab2}
\end{table*}

\subsection{Evaluation Metrics}


To comprehensively compare our proposed model with other state-of-the-art methods, we utilize four popular metrics to evaluate the COD performance. $\bullet$ Mean absolute error (MAE, $M$)~\cite{mae} evaluates the average pixel-level relative error between the normalized prediction and the ground truth. $\bullet$ Weighted F-measure ($F_\beta^w$)~\cite{fm} is an overall performance measurement that synthetically considers both weighted precision and weighted recall. $\bullet$ S-measure ($S_\alpha$)~\cite{sm} computes the object-aware and region-aware structure similarities between the prediction and the ground truth. $\bullet$ E-measure ($E_\phi$)~\cite{21Fan_HybridLoss} is based on the human visual perception mechanism to evaluate the overall and local accuracy of COD.

\subsection{Performance Comparison}
We compare the proposed method with 14 state-of-the-art COD methods, including FPN~\cite{fpn}, MaskRCNN~\cite{maskrcnn}, PSPNet~\cite{pspnet}, UNet++~\cite{unet++}, PiCANet~\cite{picanet}, MSRCNN~\cite{masksrcnn}, BASNet~\cite{basnet}, PFANet~\cite{pfanet}, CPD~\cite{cpd}, HTC~\cite{htc}, EGNet~\cite{egnet}, ANet-SRM~\cite{anet}, SINet~\cite{fan2021concealed} and MirrorNet~\cite{mirrornet}. 
For a fair comparison, the results of the first 13 methods are taken from \cite{fan2021concealed}. Besides, the results of MirrorNet based on ResNet50 are taken from \href{https://sites.google.com/view/ltnghia/research/camo}{https://sites.google.com/view/ltnghia/research/camo}.

\subsubsection{Quantitative Evaluation}

Table~\ref{tab1} summarizes the quantitative results of different COD methods on the three benchmark datasets. On the three datasets, the four evaluation indicators all have better performance than previous methods. Specifically, compared with the ResNet50-based SINet specifically designed for COD tasks, $S_\alpha$ increased by 4.54$\%$ on average, $E_\phi$ increased by 8.71$\%$ on average, and $F_\beta^w$ increased by 18.35$\%$ on average. 

\subsubsection{Qualitative Evaluation}

\begin{figure*}[ht]
    \begin{centering}
	\includegraphics[width=\textwidth]{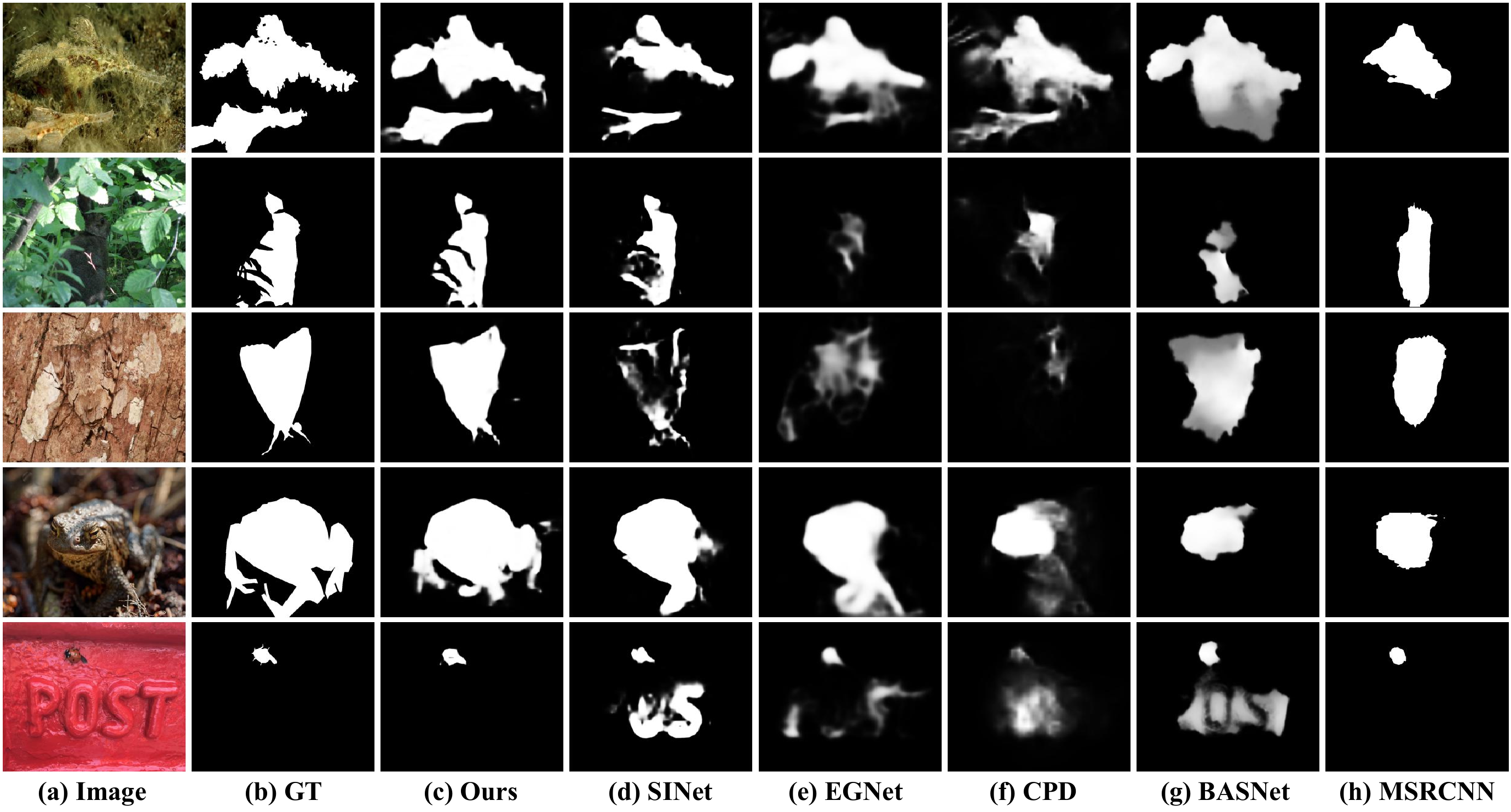}
	\caption{Qualitative results of our model and five state-of-the-art methods (\ie, 
	SINet~\protect\cite{fan2021concealed}, EGNet~\protect\cite{egnet}, CPD~\protect\cite{cpd}, BASNet~\protect\cite{basnet}, and MSRCNN~\protect\cite{masksrcnn}).
	}
    \label{results}
    \end{centering}
\end{figure*}

Figure~\ref{results} shows the qualitative results of different COD methods. Examples are from the five super-classes of COD10K from top to bottom: aquatic, terrestrial, flying, amphibious, and others. Compared with other cutting-edge models, our model can achieve better visual effects by detecting more accurate and complete camouflaged objects with rich details.

\subsection{Ablation Study}
In order to verify the effectiveness of each key module, we designed three ablation experiments, and the results are shown in Table~\ref{tab2}. In No.1 (Basic) experiment, we removed all ACFMs and DGCMs while keeping RFB modules, and then simply fused the features from the last three layers via a summation operation. In No.2 (Basic+ACFM) experiment, we removed the DGCMs and directly connected the two ACFMs. In No.3 (Basic+DGCM) experiment, we replaced the ACFMs with the combined operation of upsampling and then adding, and the DGCMs remained unchanged. No.4 (Basic+ACFM+DGCM) is the complete model, which is consistent with the structure of Figure~\ref{fig:net}. Besides, we replaced all the MSCAs in the whole network with the convolution layers to verify the effectiveness of MSCA, \emph{i.e.}, ``MSCA$\rightarrow$Conv'' in Table~\ref{tab3}.

\textbf{Effectiveness of ACFM}. We investigate the importance of the ACFM. From Table~\ref{tab2}, we observe that No.2 (Basic + ACFM) outperforms No.1 (Basic), clearly showing that the attention-induced cross-level fusion module is necessary for improving performance.

\begin{figure}[h]
    \centering{
	\includegraphics[width=1\linewidth]{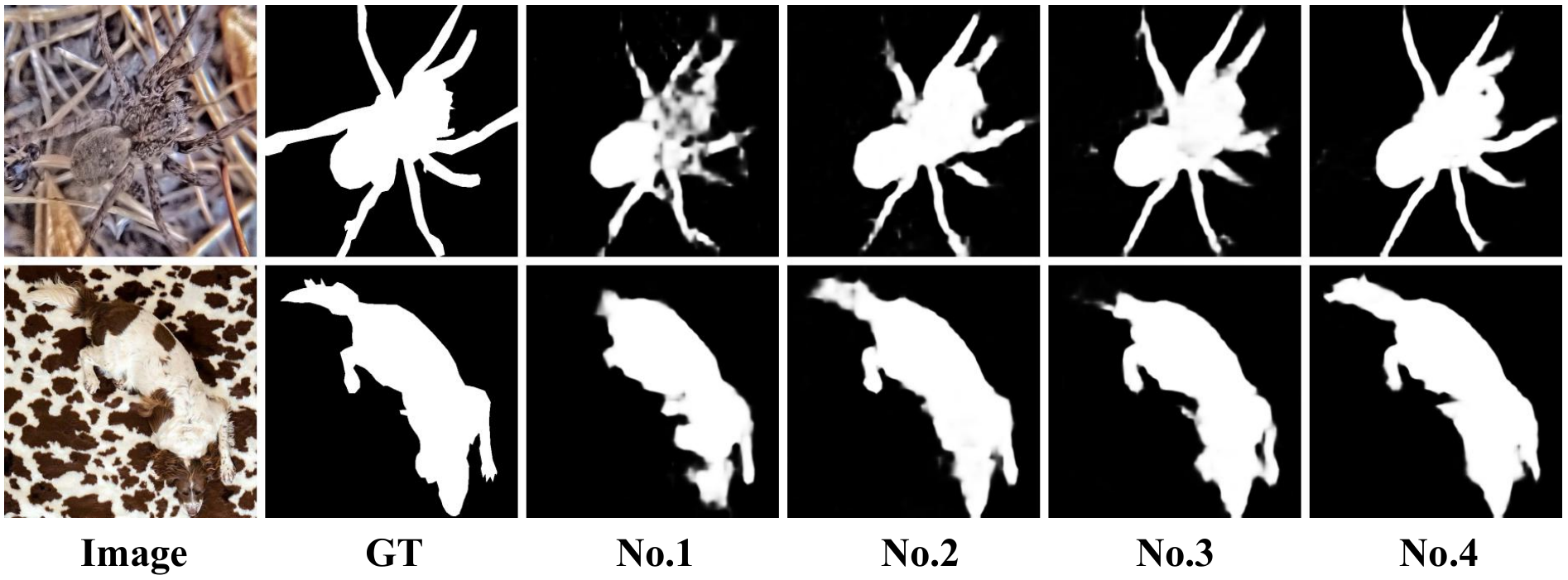}
	\caption{Visual comparisons for showing the benefits of different modules. 
	No.1, No.2, No.3 and No.4 as shown in the Table~\ref{tab2}.}
    \label{fig:ablation}
    }\vspace{-0.25cm}
\end{figure}

\textbf{Effectiveness of DGCM}. We further investigate the contribution of the DGCM. We observe that No.3 (Basic + DGCM) improves No.1 (Basic) performance on the three benchmark datasets. These improvements suggest that introducing dual-branch global context module can enable our model to accurately detect objects.

\textbf{Effectiveness of ACFM \& DGCM}. To assess the combination of the ACFM and DGCM, we evaluate the performance of No.4. As shown in Table~\ref{tab2}, our model is generally better than other settings. The visual comparison results in Figure~\ref{fig:ablation} also show that the complete structure is more conducive to detecting camouflaged objects.

\textbf{Effectiveness of MSCA}. To validate the effectiveness of MSCA, we utilize a convolution layer instead of it (denoted ``MSCA$\rightarrow$Conv''). The comparison results are shown in Table~\ref{tab3}. It can be observed that C$^2$F-Net outperforms ``MSCA$\rightarrow$Conv'' on three datasets in terms of two evaluation metrics. In particular, the use of MSCA significantly improves the $F_\beta^w$ score by a large margin of 1.9\% on the CAMO-Test dataset.

\begin{table}[th]
\caption{Ablation study on MSCA.}\vspace{-0.25cm}
\resizebox{\columnwidth}{!}{
\begin{tabular}{l|ll|ll|ll}
\hline
\multirow{2}{*}{Method} & \multicolumn{2}{l|}{CAMO-Test} & \multicolumn{2}{l|}{COD10K-Test} & \multicolumn{2}{l}{CHAMELEON} \\ \cline{2-7} 
 & $E_\phi\uparrow$ & $F_\beta^w\uparrow$ & $E_\phi\uparrow$ & $F_\beta^w\uparrow$ & $E_\phi\uparrow$ & $F_\beta^w\uparrow$ \\ \midrule
MSCA$\rightarrow$Conv  &0.843  &0.700  &0.885  &0.679  &0.934  &0.826\\
C$^2$F-Net  &\textbf{0.854}  &\textbf{0.719}  &\textbf{0.890}  &\textbf{0.686}  &\textbf{0.935}  &\textbf{0.828}\\
\hline
\end{tabular}}
\label{tab3}
\end{table}

\section{Conclusion}
In this paper, we propose a novel network \OurModel~for camouflaged object detection, which integrates the cross-level features with the consideration of rich global context information. We propose a context-aware module, DGCM, which exploits global context information from the fused features. We integrates the cross-level features with an effective fusion module, ACFM, which integrates the features with the consideration of valuable attention cues provided by MSCA. Extensive experimental results on three benchmark datasets demonstrate that the proposed model outperforms other state-of-the-art methods. 


{
\bibliographystyle{named}
\bibliography{ijcai21}
}

\end{document}